\documentclass[sigconf,nonacm]{acmart}
\setcopyright{none}

\AtBeginDocument{%
  }

\usepackage[textsize=tiny]{todonotes}

\usepackage[table]{xcolor}  
\usepackage{colortbl}       
\usepackage{xurl}
\usepackage{hyperref}
\usepackage{graphicx}

\usepackage{fontawesome5}

\usepackage{fontawesome5}
\definecolor{ModelGreen}{RGB}{213,232,212}
\newcolumntype{L}[1]{>{\raggedright\arraybackslash}p{#1}}
\begin{document}

\title{CoME-VL: Scaling Complementary Multi-Encoder Vision-Language Learning}

\author{Ankan Deria*, Komal Kumar*, Xilin He, Imran Razzak, Hisham Cholakkal, Fahad Shahbaz Khan, Salman Khan}
\affiliation{
  \institution{\textbf{Mohamed bin Zayed University of Artificial Intelligence}}
  \city{Abu Dhabi}
  \country{UAE}
}

\affiliation{ 
  \institution{%
    \parbox{0.9\linewidth}{\centering
    \vspace*{10pt}
    \small
    \raisebox{-0.15em}{\includegraphics[height=0.9em]{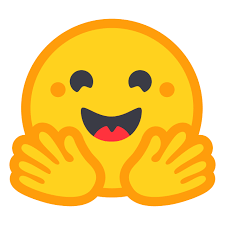}}~%
    \textbf{Model:}~
    \href{https://huggingface.co/MBZUAI/CoME-VL}{\textcolor{teal}{\url{huggingface.co/MBZUAI/CoME-VL}}}
    \\[2pt]
    {\fontsize{10}{10}\selectfont\faGithub}~\textbf{GitHub:}~
    \href{https://github.com/mbzuai-oryx/CoME-VL}{\textcolor{teal}{\url{github.com/mbzuai-oryx/CoME-VL}}}
    \\[2pt]
    {\fontsize{10}{10}\selectfont\faGlobe}~\textbf{Project Page:}~
    \href{https://mbzuai-oryx.github.io/CoME-VL/}{\textcolor{teal}{\url{mbzuai-oryx.github.io/CoME-VL}}}
    }
  }
  \city{}
  \country{}
}


\renewcommand{\shortauthors}{}


\begin{abstract}
  Recent vision-language models (VLMs) typically rely on a single vision encoder trained with contrastive image-text objectives, such as CLIP-style pretraining. While contrastive encoders are effective for cross-modal alignment and retrieval, self-supervised visual encoders often capture richer dense semantics and exhibit stronger robustness on recognition and understanding tasks. In this work, we investigate how to scale the fusion of these complementary visual representations for vision-language modeling.
We propose \textbf{CoME-VL} (\textbf{C}omplementary \textbf{M}ulti-\textbf{E}ncoder \textbf{V}ision-\textbf{L}anguage), a modular fusion framework that integrates a contrastively trained vision encoder with a self-supervised DINO encoder. Our approach performs representation-level fusion by (i) entropy-guided multi-layer aggregation with orthogonality-constrained projections to reduce redundancy, and (ii) RoPE-enhanced cross-attention to align heterogeneous token grids and produce compact fused visual tokens. The fused tokens can be injected into a decoder-only LLM with minimal changes to standard VLM pipelines.
Extensive experiments across diverse vision-language benchmarks demonstrate that CoME-VL consistently outperforms single-encoder baselines. In particular, we observe an average improvement of \textbf{4.9\%} on visual understanding tasks and \textbf{5.4\%} on grounding tasks. Our method achieves state-of-the-art performance on RefCOCO for detection while improving over the baseline by a large margin. Finally, we conduct ablation studies on layer merging, non-redundant feature mixing, and fusion capacity to evaluate how complementary contrastive and self-supervised signals affect VLM performance. We will release our code as open-source on GitHub.
\end{abstract}

\keywords{Vision-language models, multi-encoder fusion, multimodal understanding, entropy-guided layer selection, orthogonal layer, RoPE}
\begin{teaserfigure}
\centering
  \includegraphics[width=\linewidth]{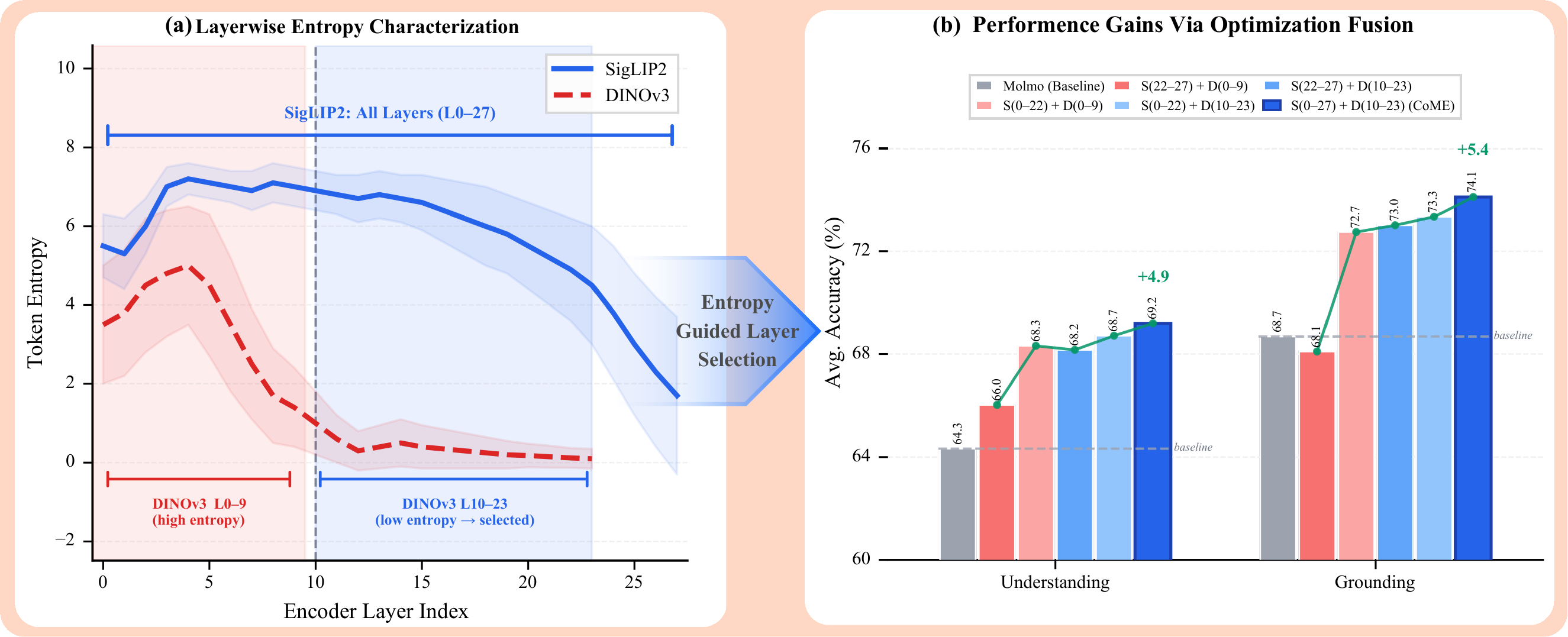}
  \captionof{figure}{CoME-VL uses token entropy analysis to identify complementary layer ranges from multiple vision encoders (SigLIP2 and DINOv3). By composing all SigLIP2 layers (which exhibit high entropy, capturing diverse semantic features) with the low-entropy DINOv3 layers 10--23 (which encode strong spatial features), CoME-VL achieves consistent improvements over the Molmo~\cite{deitke2025molmo} baseline (single-encoder), averaging +4.9\% on visual understanding/generation and +5.4\% on grounding tasks.}
  \label{fig:teaser}
\end{teaserfigure}


\maketitle
\settopmatter{printfolios=true}
\pagestyle{plain}

\section{Introduction}\label{sec:intro}
\label{sec:intro}
Large Language Models (LLMs)~\cite{chatgpt, openai2023gpt4, touvron2023llama, llama2, taori2023stanford, chiangvicuna} have achieved major progress in language understanding and generation. Building on these advances, vision--language models~\citep{li2023blip,Qwen-VL,wang2023visionllm,chen2024_595,laurenon2024_598,yang2024_637, li2023otter,liu2023internchat,liu2023llava,alayrac2022flamingo, deria2024inverge} extend LLMs to multimodal inputs such as images and videos, enabling tasks including visual question answering, image captioning, and instruction-following dialogue grounded in visual content. A key enabler of this progress is instruction tuning~\cite{wei2021finetuned,wang2022self}, which turns pretrained language models into general-purpose systems that can adapt to diverse tasks through natural language prompts, often with strong zero-shot transfer. These capabilities motivate integrating visual signals into LLMs to keep generated text grounded in visual observations.

Early multimodal systems such as Flamingo~\cite{alayrac2022flamingo} and BLIP2~\cite{li2023blip2} align a frozen visual encoder with a powerful LLM to handle a variety of vision--language tasks. Subsequent work further improves instruction following by constructing multimodal instruction datasets and tuning the joint model, including LLaVA~\cite{liu2023llava}, InstructBLIP~\cite{dai2023instructblip}, MiniGPT-4~\cite{zhu2023minigpt}, and mPLUG-OWL~\cite{ye2023mplug}. However, many of these approaches rely primarily on image-level alignment, which can limit fine-grained understanding needed for region-level description~\cite{liu2017referring}, compositional reasoning~\cite{zellers2019recognition}, and can exacerbate object hallucination~\cite{li2023obj_Hallucination}. To address this limitation, recent methods introduce region supervision and grounding: GPT4RoI~\cite{zhang2023gpt4roi} performs region-of-interest instruction tuning, while Kosmos-2~\cite{peng2023kosmos} and Shikra~\cite{chen2023shikra} integrate referring and grounding into dialogue, enabling models to respond with spatial coordinates for mentioned objects. Such grounding capabilities substantially expand the range of vision--language applications.

Despite these advances, most multimodal LLMs still adopt CLIP~\cite{radford2021learning} or CLIP-style variants~\cite{sun2023eva} as the visual encoder, typically using a single deep-layer representation (often the penultimate layer) as input to the language model. This common practice raises an important open question: \emph{Is using vanilla CLIP features the best visual representation for multimodal LLMs?} Although CLIP is strongly aligned with text through image--text contrastive learning, its supervision is largely global and can underemphasize fine-grained visual cues such as precise localization, color attributes, and subtle spatial relationships. In addition, current MLLMs are often architecturally imbalanced, pairing comparatively smaller vision encoders (e.g., ViT-Large) with much larger language backbones (e.g., 7B--13B LLMs). As language models continue to benefit from scaling \cite{kumar2025llm}, the visual encoder can become the bottleneck in the overall system, limiting generalization, worsening domain transfer, and restricting the emergence of higher-level multimodal abilities. Therefore, improving the visual representation and its integration strategy is critical for advancing grounded and reliable multimodal LLMs.

Prior works~\citep{jiang2023clip, li2024eagle,tong2024cambrian} systematically study visual encoders for multimodal LLMs and show that combining features from different encoders can improve downstream performance. However, effectively leveraging multiple encoders remains challenging: naive fusion can introduce redundant features and substantially increase the number of visual tokens processed by the LLM.

To address this, we propose \textbf{CoME-VL} (Complementary Multi-Encoder Vision-Language), a principled multi-encoder fusion framework that combines SigLIP and DINO to capture complementary semantic and grounding information for multimodal learning. CoME-VL incorporates \emph{entropy-guided layer selection} to identify informative features across encoder depths, \emph{orthogonality-regularized fusion} to suppress redundancy and preserve complementary representations, and \emph{RoPE-enhanced cross-attention} to efficiently align and merge encoder features without directly concatenating visual patches, thus avoiding unnecessary burden on the LLM. Our analysis shows that SigLIP contributes strong semantic features for visual understanding, whereas DINO provides richer grounding cues, and that the best performance is obtained by combining DINO's late-layer features with SigLIP features aggregated across layers. Extensive experiments demonstrate consistent improvements over strong baselines, establishing CoME-VL as an effective and efficient framework for grounded multimodal learning.



The contributions of this paper are summarized as follows:
\begin{itemize}
\item We propose \textbf{CoME-VL}, a complementary multi-encoder vision-language framework that integrates representations from a self-supervised vision encoder and a contrastive vision-language encoder to improve both visual understanding and grounding.
\item We introduce \textbf{layer-wise entropy} as a principled criterion for selecting informative layers across encoder depths and guiding effective multi-layer feature fusion.
\item We design an \textbf{orthogonality-regularized fusion module} to reduce redundant information across multiple encoders, together with \textbf{RoPE-enhanced cross-attention} to align heterogeneous features efficiently without directly increasing the visual token burden on the LLM.
\item We conduct comprehensive experiments and analyses showing that our approach consistently improves grounding and understanding performance over strong baselines while maintaining efficient inference.
\end{itemize}



\begin{table*}[!ht]
    \centering
    \caption{Comparison of recent vision-language models on PixMo counting and pointing tasks. Single-encoder models (CLIP/SigLIP-based) show limited grounding performance, while the proposed multi-encoder CoME-VL (DINO + SigLIP) achieves improved accuracy. - denotes not supported.}

    \label{tab:single_enc_comp}
    \resizebox{\textwidth}{!}{
    \begin{tabular}{lccccccc}
    \toprule
        Model & MiniCPM – 4.5 & Intern-VL 3.5 4B & Intern-VL 3.5 8B & QWEN3-VL 4B & QWEN3-VL 8B & molmo 7B & Dino+Siglip(CoME-VL 7B) \\ 
        \midrule
        Chart & 54.98 & 50.48 & 55.56 & 57.71 & 43.46 & 52.39 & 57.24 \\ 
        Counting & 64.52 & 83.8 & 83.8 & 78.8 & 69.04 & 83.31 & 87.83 \\
        Pointing (@3px/5px) & - & - & - & - & - & 53.79/68.94 & 58.56/75.94 \\ 
        \bottomrule
    \end{tabular}
    }
    \vspace{-5 pt}
\end{table*}


\section{Related Work}

\subsection{Vision-Language foundation models}
Recent progress in vision-language modelling has been driven by aligning strong language backbones with pretrained visual representations, enabling general-purpose multimodal understanding and generation. A representative direction is to train or adapt lightweight visual-to-language interfaces while keeping large components fixed or minimally tuned. For example, OpenFlamingo~\cite{awadalla2023_600,singh2022flava} provides an open framework for training autoregressive vision-language models, while instruction-tuned models such as InstructBLIP~\cite{dai2023_584} and MiniGPT-4~\cite{zhu2023_588} demonstrate that instruction supervision can substantially improve usability across diverse vision-language tasks. More recent work continues to expand the modelling space to improve reasoning and efficiency, including chain-of-thought style visual reasoning~\cite{xu2024_605}, mixture-of-experts designs~\cite{lin2024_609}, and distillation or synthesized supervision to build lighter yet capable models~\cite{chen2024_622}. Some works on image retrieval~\cite{koh2023grounding}, video understanding~\cite{zhang2023video}, biomedical analysis~\cite{li2023llava, deria2026medmo,deria2024inverge}, control systems~\cite{driess2023palme} also extend the vlm. In parallel, scaling and improving contrastive vision-language pretraining has produced stronger visual branches that serve as backbones for downstream VLMs, such as EVA-CLIP and its larger variants~\cite{sun2023_293,sun2024_311}.

\subsection{Multi-encoder visual features for VLMs}
Despite the diversity of MLLM architectures, many systems still rely on a single visual encoder representation, often taken from the last or penultimate layer of a CLIP-style model, which biases the input toward global semantics and can underrepresent localized cues. Jiang et al.~\cite{jiang2023clip} analyze this issue by comparing CLIP-style image-text contrastive features and DINO-style self-supervised features in multimodal LLMs, and study multi-stage feature merging to enhance representations. Following this insight, several works explore combining complementary encoders or transferring useful properties between them. For instance, CLIP meets DINO~\cite{imam2024_61} and CLIP-DINOiser~\cite{wysoczanska2023_66} study how DINO-like signals can improve or adapt CLIP representations, while Frozen CLIP-DINO~\cite{zhang2025_10} highlights the strength of hybrid CLIP-DINO backbones in downstream dense prediction. Additional evidence suggests that the choice of supervision and encoder type measurably affects downstream performance and robustness, motivating careful encoder selection and fusion for multimodal models~\cite{liu2025_31}. Surveys of multimodal fusion also emphasize that encoder complementarity and fusion design are central to improving vision-language systems beyond single-encoder pipelines~\cite{han2025_635}. These findings collectively motivate multi-encoder and multi-layer feature utilization as a practical path toward stronger fine-grained perception and grounding in VLMs. Many works study multi-encoder feature merging~\cite{shi2024eagle, li2024eagle, tong2024cambrian, jiang2023clip, karamcheti2024prismatic}, and their analysis suggests that DINO and SigLIP form an effective combination; however, this has not been evaluated on grounding tasks, which require both semantic and spatial information. Recent depth-aware fusion strategies in MLLMs, such as depth--breadth fusion in Florence-VL~\cite{chen2025florence}, together with findings that large-scale language-free/self-supervised visual representations can approach language-supervised contrastive encoders~\cite{fan2025scaling}, further support the use of DINO-like and SigLIP-like features as complementary. Our work takes a step further by analyzing complementary feature merging from encoder-level to layer-wise fusion, focusing on DINO and SigLIP. Our analysis suggests that this combination is not only suited for VQG and captioning but also effective for grounding tasks such as pointing, counting, and bounding box prediction. Finally, while alignment mechanisms have been explored in downstream VL tasks such as navigation (DELAN~\cite{du2024delan}) and concept modeling with instruction tuning (VCM~\cite{luo2025vcm}), our contribution is complementary: we propose an entropy-guided fusion analysis that informs encoder- and layer-wise feature merging.

\section{Preliminary Analysis}\label{sec:motivation}


Before presenting our methodology, we conduct a series of preliminary studies to understand the limitations of existing approaches and motivate our design choices. These analyses reveal important insights that guide the design of our multi-encoder framework.

\paragraph{\textbf{Single-Encoder Architectures.}}
Most existing vision-language models (VLMs) rely on a single vision encoder~\citep{deitke2025molmo, wang2024qwen2, wang2025internvl3, yu2025minicpm}, typically CLIP~\cite{radford2021learning} or SigLIP~\cite{zhai2023sigmoid}, to extract visual features before projecting them into the language model's embedding space. While this design has proven effective for general image understanding, we observe systematic failures in fine-grained visual grounding tasks such as counting and pointing. Table~\ref{tab:single_enc_comp} shows that recent CLIP-style single-encoder models perform poorly on fine-grained grounding tasks such as pointing. While these works~\cite{jiang2023clip, shi2024eagle, tong2024cambrian, karamcheti2024prismatic} suggest in their analysis that multi-encoders (SAM, MAE, Siglip, CLIP, DINO, ConvNEXT) improve the performance on understanding scenarios in which DINO does not help much in the combination with others, these methods were not tested on the tasks that require visual grounding. COMM~\cite{jiang2023clip} investigates that DINO provides the complementary features that improve these tasks.

\begin{figure*}[t]
    \centering
    \includegraphics[width=0.96\linewidth]{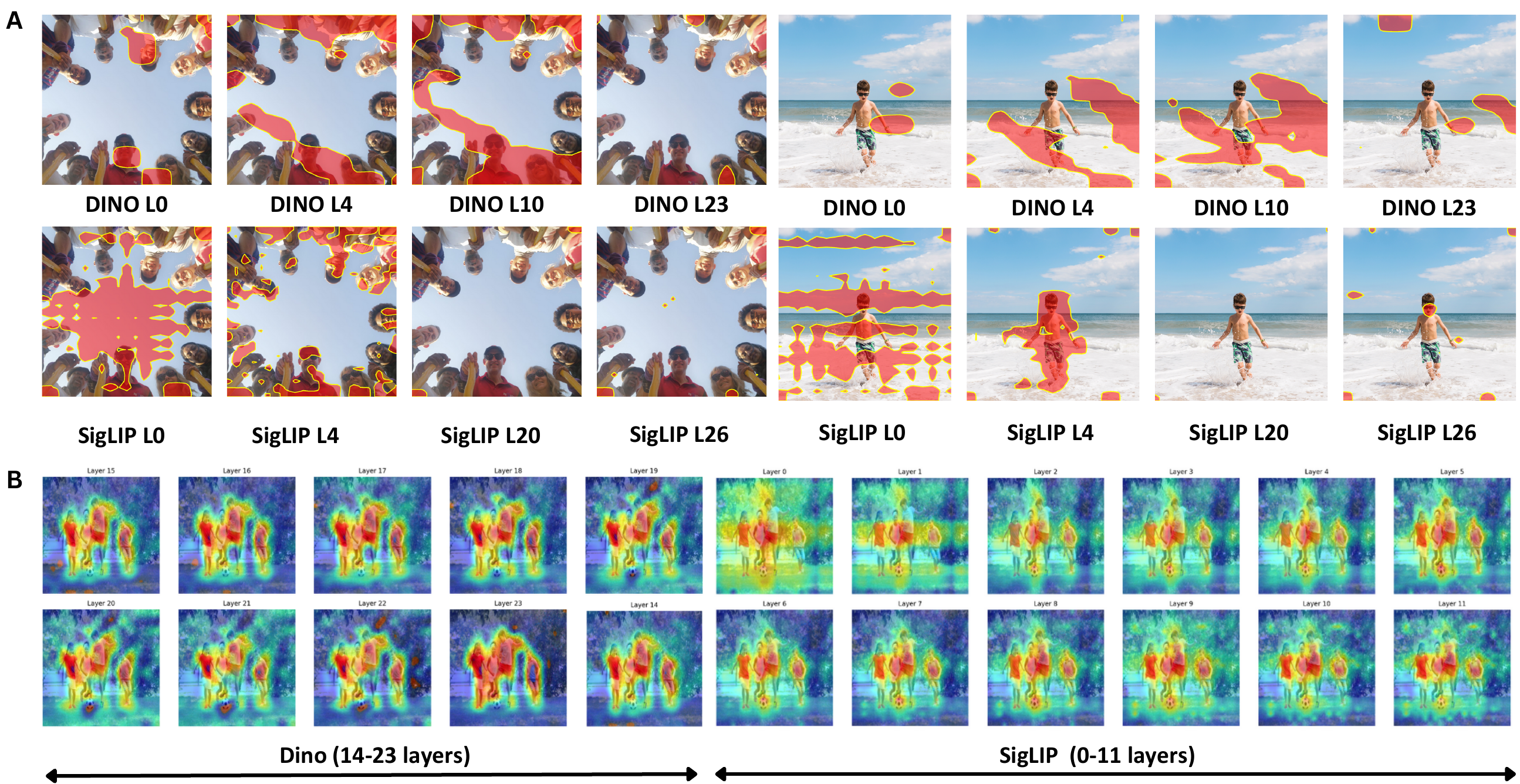}
    \caption{Semantic feature analysis. (a) Layer-wise comparison of spatial attention in DINOv3 and SigLIP2. We visualize attention masks from four representative layers per model (early, early-mid, mid-late, final), showing the top 30\% attention with contour overlays. (b) Layer-wise attention rollout visualization using Gard-CAM for the selective layers from Fig \ref{fig:teaser}. (Best view in zoom)}
    \label{fig:sementic}
    
\end{figure*}

\noindent\textbf{Semantic-Visual Trade-off.}
Figure~\ref{fig:sementic} shows a clear difference in how spatial attention evolves in DINOv3 and SigLIP2 across depth. In the top-row attention masks (Fig.~\ref{fig:sementic}A), DINOv3 produces spatially coherent and object-centric patterns already from early layers, and this behavior persists as features become more structured across depth, indicating a strong bias toward geometry, boundaries, and localized regions. In contrast, SigLIP2 exhibits more diffuse and fragmented attention in shallow layers and only gradually concentrates on a few semantically informative regions in deeper layers, reflecting its global image--text alignment objective. The layer-wise rollout visualizations (Fig.~\ref{fig:sementic}B) further support this observation: DINOv3 emphasizes the spatial extent of objects, while SigLIP2 increasingly highlights discriminative parts that are useful for semantic matching rather than precise localization.

These complementary properties (Table~\ref{tab:single_enc_comp} and Fig.~\ref{fig:sementic}) motivate our design choice of \textbf{dual encoders}, combining SigLIP2 for semantic alignment with DINOv3 for fine-grained spatial cues.

\paragraph{\textbf{Spatial Entropy Analysis.}}
Fig.~\ref{fig:teaser}(b) shows that spatial entropy decreases with depth, reflecting a shift from broadly distributed low-level evidence to concentrated high-level activations. This behavior matches Fig.~\ref{fig:sementic} A, where SigLIP2 maintains more object-centric spatial attention while DINOv3 gradually focuses on semantically discriminative regions. These complementary trends motivate multi-layer fusion for localization cues and semantic alignment.

\paragraph{\textbf{Redundancy in Multi-layer Fusion.}}
Naïvely concatenating or averaging features from many layers often introduces redundancy, since neighboring transformer layers tend to encode highly similar information. Fig.~\ref{fig:sementic} B (Also See the Appendix~\ref{apd:sementic} for details) supports this by showing strong inter-layer similarity and a concentrated subspace in the fused representation, indicating that simple fusion wastes representational capacity. This motivates our \textbf{orthogonality-layer mixing}, which explicitly encourages complementary and decorrelated features before fusion.

\paragraph{\textbf{Fusion and Alignment in Multi-encoder Fusion.}}
Fusion has been actively explored in recent vision-language models. Many MLLMs, such as LLaVA~\cite{liu2023llava} and Qwen-VL~\cite{Qwen-VL}, fuse visual tokens and text tokens by concatenation followed by self-attention in the language backbone. While effective, this design increases the number of tokens processed by the LLM and can lead to higher compute and memory cost. We therefore adopt a cross-attention-based interface, in the spirit of BLIP-2~\cite{li2023blip}, where the language tokens attend to a compact set of visual tokens, providing a more efficient and controllable fusion mechanism. Please see Appendix~\ref{app:fusion_alignment}.

Multi-encoder fusion further introduces a spatial alignment challenge because different encoders often output token grids at different resolutions. Naïve cross-attention between mismatched grids can overemphasize semantic similarity and attend to spatially distant regions, which is undesirable for precise grounding. To address this, we incorporate \textbf{RoPE-based cross-attention alignment}, which injects relative positional structure into the attention computation, encouraging spatially coherent interactions while preserving the ability to use global context.

\section{Methodology}\label{sec:method}

\begin{figure*}[ht]
    \centering
    \includegraphics[width=0.8\linewidth]{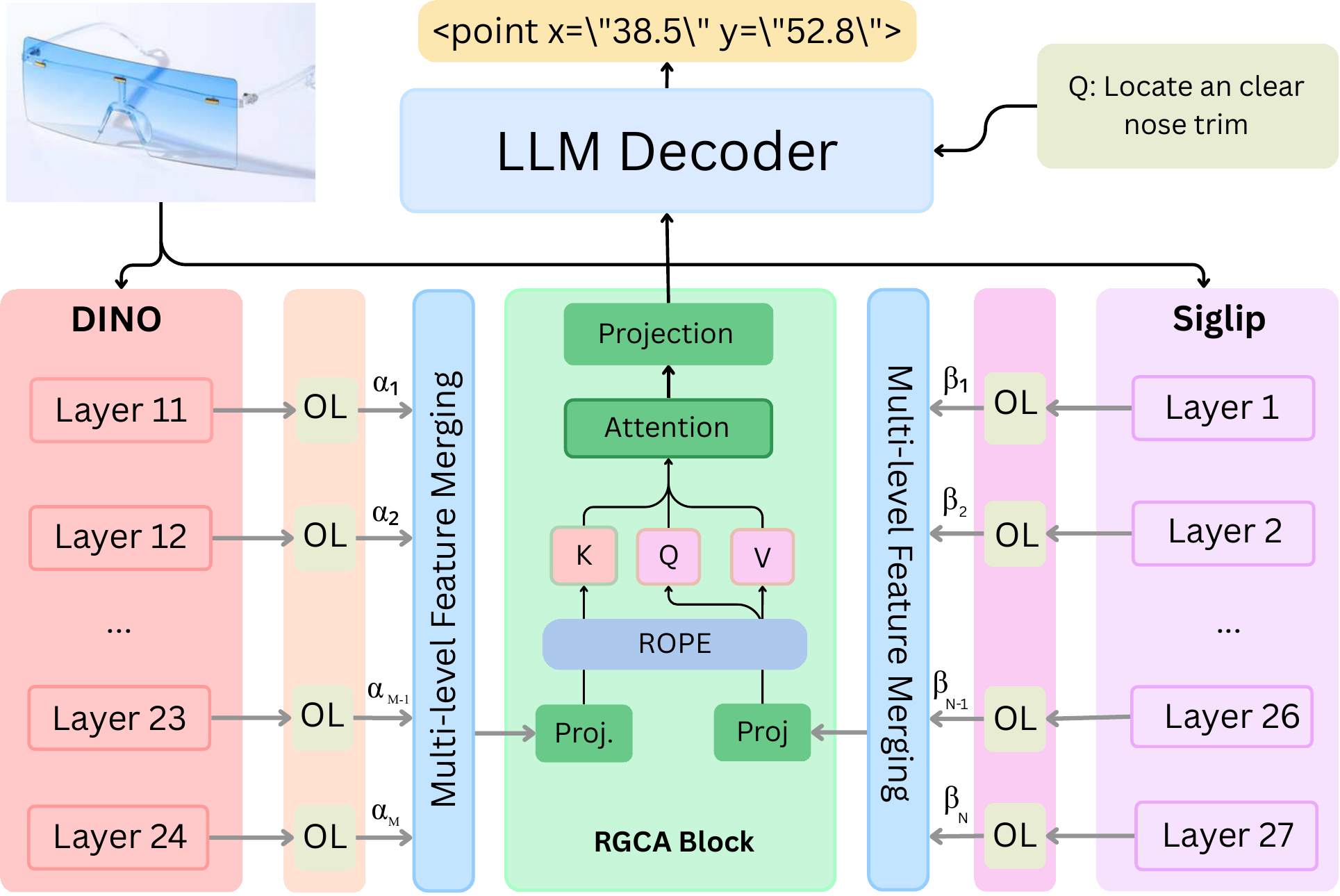}
    \caption{Overview of the proposed multi-encoder, multi-scale vision-language framework. Images are processed by two complementary vision encoders (SigLIP2 and DINOv3), with hierarchical features extracted across layers, fused via orthogonality-regularized mixing (OL), spatially aligned using RoPE-based cross-attention, and injected into a decoder-only language model for grounding and generation.}
    \label{fig:main_dig}
    
\end{figure*}

Building on our preliminary analysis, we propose a multi-encoder, multi-layer alignment framework for visual grounding. As shown in Fig.~\ref{fig:main_dig}, our architecture comprises three key components: (1) \textbf{dual complementary visual encoders} that provide semantic alignment and fine-grained spatial cues, (2) \textbf{learnable multi-layer aggregation} with an \textbf{orthogonality-regularized mixer} to reduce redundancy across layers, and (3) \textbf{RoPE-enhanced cross-attention} that aligns tokens from heterogeneous encoders and injects the fused visual tokens into a decoder-only language model.

\subsection{Multi-layer Visual Representations}
Motivated by the semantic-spatial trade-off observed in Fig.~\ref{fig:sementic}, we extract hierarchical representations from two complementary encoders, SigLIP2 and DINOv3, and aggregate features across depth rather than relying on a single late-layer representation.

\paragraph{\textbf{SigLIP2 feature hierarchy.}}
SigLIP2 is a vision transformer pretrained with a vision-language contrastive objective, producing representations that are strongly aligned with text. Given an image $I \in \mathbb{R}^{H \times W \times 3}$, we tokenize it into $N_s=\frac{H}{P}\times\frac{W}{P}$ patches of size $P\times P$ and embed them into $d_s$-dimensional tokens. Let $\mathbf{Z}^{(l)}_{\text{Sig}} \in \mathbb{R}^{N_s \times d_s}$ denote the hidden states at layer $l$ of SigLIP2:
\begin{equation}
\mathbf{Z}^{(l)}_{\text{Sig}} = \mathrm{Layer}^{(l)}_{\text{Sig}}(\mathbf{Z}^{(l-1)}_{\text{Sig}}), \quad l=1,\ldots,27.
\end{equation}
Following Fig.~\ref{fig:main_dig}, we retain the intermediate outputs from layers $\mathcal{L}_s=\{1,\ldots,27\}$.

\paragraph{\textbf{DINOv3 feature hierarchy.}}
DINOv3 is trained with self-supervised objectives and captures fine-grained structure such as boundaries and part-level geometry. Let $\mathbf{Z}^{(l)}_{\text{DINO}} \in \mathbb{R}^{N_d \times d_d}$ denote the DINOv3 hidden states at layer $l$. Following Fig.~\ref{fig:main_dig}, we use the subset $\mathcal{L}_d=\{11,\ldots,24\}$, which is guided by our entropy analysis and per-layer grounding performance (Sec.~\ref{sec:motivation}).


\paragraph{\textbf{Entropy-informed candidate layer selection.}}
Layer-wise entropy reveals distinct depth-wise behavior in SigLIP2 and DINOv3: earlier layers tend to be more spatially diffuse, whereas deeper layers become increasingly concentrated on salient regions. We use this analysis to define compact candidate layer sets, denoted by $\mathcal{L}_s$ for SigLIP2 and $\mathcal{L}_d$ for DINOv3. Compared with arbitrary depth thresholds or random subset selection, entropy provides a data-dependent signal for restricting the fusion space according to encoder-specific depth behavior. Given these candidate sets, we then learn soft aggregation weights end-to-end:

\begin{equation}
\begin{aligned}
w_l^{(s)} &= \frac{\exp(\alpha_l)}{\sum_{k \in \mathcal{L}_s} \exp(\alpha_k)}, \quad l \in \mathcal{L}_s,\\
w_l^{(d)} &= \frac{\exp(\beta_l)}{\sum_{k \in \mathcal{L}_d} \exp(\beta_k)}, \quad l \in \mathcal{L}_d.
\end{aligned}
\end{equation}
The resulting weighted fusion allows the model to adaptively combine complementary information across depth, including broader contextual semantics and finer spatial cues.

\begin{table*}[htbp]
\centering
\caption{Performance comparison of vision-language models on PixMo benchmarks. CoME-VL consistently improves visual understanding and grounding, with notable gains on counting and pointing tasks (@3px/5px). - denotes not supported. }
\small
\setlength{\tabcolsep}{6pt}
\renewcommand{\arraystretch}{1.2}

\resizebox{0.87\textwidth}{!}{%
\begin{tabular}{
lcccccc
}
\toprule
\textbf{Model} &
\textbf{Chart} &
\textbf{Diagrams} &
\textbf{Tables} &
\textbf{Others} &
\textbf{Counting} &
\textbf{Pointing} \\
\midrule

LLAVA1.5-7B~\cite{liu2023llava}       & 20.31 & 28.32 & 20.80 & 29.68 & 78.27 & - \\
LLAVA1.5-13B~\cite{liu2023llava}       & 23.33 & 33.59 & 23.24 & 34.86 & 70.95 & - \\
LLAVA-Mistral 7B~\cite{liu2023llava}   & 22.16 & 33.98 & 26.95 & 48.14 & 77.82 & - \\
Intern-VL2 8B~\cite{wang2025internvl3}      & 57.71 & 72.85 & 73.82 & 90.62 & 74.05 & - \\
QWEN2-VL 7B~\cite{wang2024qwen2}       & 45.21 & 64.25 & 61.13 & 86.91 & 57.42 & - \\
Pixtral-12B~\cite{agrawal2024pixtral}       & 38.28 & 54.00 & 63.96 & 64.94 & 71.66 & - \\

Paligemma-3B~\cite{beyer2024paligemma}       & 16.50 & 26.26 & 20.80 & 20.11 & 8.57 & - \\
Kosmos-2 8B~\cite{peng2023kosmos}         & 7.81 & 8.88 & 12.50 & 8.00 & 26.19 & - \\
Instruct-BLIP 7B~\cite{dai2023instructblip}       & 13.28 & 17.08 & 16.60 & 10.54 & 36.19 & - \\
Phi3 7B~\cite{abdin2025phi}        & 10.54 & 9.37 & 9.57 & 7.22 & 12.61 & - \\
GLM-4V 9B~\cite{hong2025glm}       & 40.23 & 58.65 & 54.12 & 84.37 & 84.76 & - \\
Molmo2 7B~\cite{deitke2025molmo}
& 52.39 & 62.41 & 66.25 & 76.26 & 83.31 & 53.79 / 68.94 \\


\rowcolor{ModelGreen}\textbf{CoME-VL 7B}            & \textbf{57.24}    & \textbf{66.94}    & \textbf{70.75}    & \textbf{81.84}    & \textbf{87.83}    & \textbf{58.56/75.94} \\

\bottomrule
\end{tabular}%
}

\label{tab:main_table}
\end{table*}

\subsection{Orthogonality-Regularized Multi-layer Mixing}
Naive fusion across many layers can be inefficient because intermediate representations are often highly correlated across depth. To mitigate this effect, we project each selected layer through a lightweight per-layer transformation before aggregation. We instantiate this transformation as an Orthogonal Layer (OL), which preserves feature scale through a near-isometric mapping while encouraging less redundant layer mixing.

\paragraph{\textbf{Orthogonal Layer (OL)}}
As shown in Fig.~\ref{fig:main_dig}, we apply an Orthogonal Layer (OL) to each selected layer output before aggregation. OL is a lightweight linear projection $\mathbf{z}_i = \mathbf{Q}_i \mathbf{h}_i$ whose weight $\mathbf{Q}_i \in \mathbb{R}^{m \times d}$ is constrained to be orthogonal using such that:
\begin{equation}
    \begin{cases}
        \mathbf{Q}_i^{\top}\mathbf{Q}_i = \mathbf{I}_d, & \text{if } m \geq d, \\
        \mathbf{Q}_i\mathbf{Q}_i^{\top} = \mathbf{I}_m, & \text{if } m < d.
    \end{cases}
\end{equation}
For square matrices ($m = d$), we parameterize $\mathbf{Q}_i$ through a learned skew-symmetric matrix $\mathbf{A}_i \in \mathbb{R}^{d \times d}$ (where $\mathbf{A}_i = -\mathbf{A}_i^{\top}$) via $\mathbf{Q}_i = \exp(\mathbf{A}_i)$ or the Cayley transform $\mathbf{Q}_i = (\mathbf{I}_d + \mathbf{A}_i/2)(\mathbf{I}_d - \mathbf{A}_i/2)^{-1}$. This promotes a near-isometric transformation, reduces redundant directions across layers, and improves optimization stability when learning the layer weights. We provide more details in Appendix~\ref{append:OL}.

\paragraph{\textbf{Intra-encoder multi-layer aggregation.}}
For each encoder $e \in \{\text{Sig},\text{DINO}\}$, we first normalize each layer output, project it with OL, and then take a weighted sum across the selected set:
\begin{equation}
\tilde{\mathbf{Z}}^{(l)}_{e} = \mathrm{OL}_{e}\!\left(\mathrm{LN}(\mathbf{Z}^{(l)}_{e})\right),
\qquad
\mathbf{V}_{e} = \sum_{l \in \mathcal{L}_{e}} w_l^{(e)} \tilde{\mathbf{Z}}^{(l)}_{e}.
\end{equation}
Here, $\mathbf{V}_{\text{Sig}} \in \mathbb{R}^{N_s \times d}$ and $\mathbf{V}_{\text{DINO}} \in \mathbb{R}^{N_d \times d}$ are pooled token sets in a shared hidden dimension $d$ (implemented by choosing OL output dimensions to match).

\subsection{RoPE-Enhanced Alignment}
Fusing heterogeneous encoders is challenging because their token grids can differ in resolution. SigLIP2 typically produces $N_s$ tokens (e.g., $24\times24$ on $384\times384$), while DINOv3 produces $N_d$ tokens (e.g., $14\times14$ depending on preprocessing). Direct token concatenation is a simple fusion strategy, but it increases the number of visual tokens processed by the LLM and therefore raises compute and memory cost. In contrast, our RoPE-enhanced cross-attention aligns heterogeneous encoder features more efficiently while preserving spatial correspondence; as shown in Table~\ref{tab:siglip_dino_ablation}, CoME-VL increases inference time only modestly from 1.26s to 1.52s per sample, and remains substantially more efficient than COMM~\cite{jiang2023clip}, which requires about 2.2s/sample due to direct feature concatenation.

\paragraph{\textbf{Geometry-aware cross-attention with RoPE.}}
We use SigLIP2 tokens as queries and DINOv3 tokens as keys and values. To encourage spatially consistent matching, we incorporate RoPE into the attention computation so that attention scores depend on relative spatial offsets:
\begin{align}
\mathbf{Q} &= \mathrm{RoPE}(\mathbf{W}_Q\, \mathrm{LN}(\mathbf{V}_{\text{Sig}})), \\
\mathbf{K} &= \mathrm{RoPE}(\mathbf{W}_K\, \mathrm{LN}(\mathbf{V}_{\text{DINO}})), \\
\mathbf{V} &= \mathbf{W}_V\, \mathrm{LN}(\mathbf{V}_{\text{DINO}}),
\end{align}
followed by standard scaled dot-product attention. The cross-attention output is combined with SigLIP2 features via a gated residual connection. We follow Flamingo~\cite{alayrac2022flamingo} for gated attention in visual fusion:
\begin{equation}
\mathbf{V}_{\text{fused}} = \mathbf{V}_{\text{Sig}} + \tanh(\gamma)\cdot \mathrm{CrossAttn}(\mathbf{V}_{\text{Sig}}, \mathbf{V}_{\text{DINO}}),
\end{equation}
where $\gamma$ is initialized to zero to stabilize early training and gradually enables DINOv3 information to be incorporated.

Finally, the projected visual tokens produced by the RGCA cross-attention block (Fig.~\ref{fig:main_dig}) are forwarded to the decoder-only LLM. Compared with concatenation followed by self-attention, this cross-attention interface injects a compact set of visual tokens and avoids increasing the LLM context length, making the fusion more compute- and memory-efficient while preserving the aligned visual evidence needed for generation and grounding. For bounding-box generation, our baseline Molmo~\cite{deitke2025molmo} does not natively support object detection. We therefore redesign the fine-tuning output format to enable box prediction (see Fig.~\ref{fig:main_dig} for the target format). For more details see Appendix~\ref{app:box_tokens}.

\begin{figure*}[t]
    \centering
    \includegraphics[width=\linewidth]{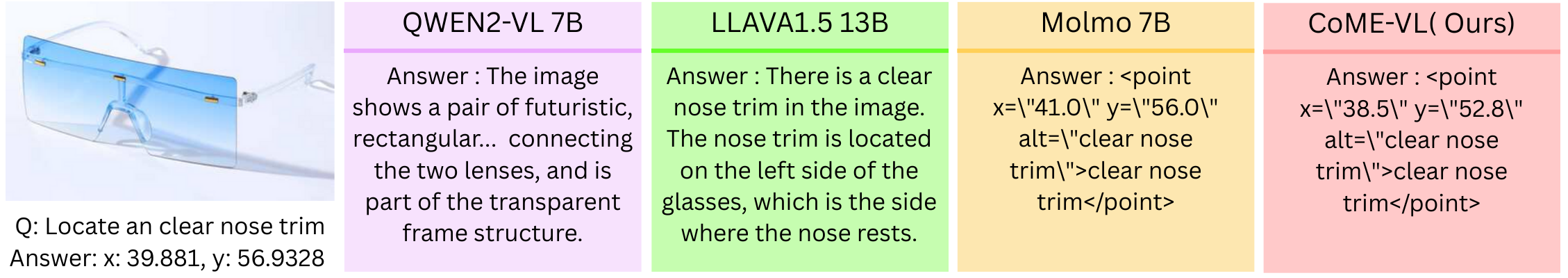}
    \caption{Qualitative results on PixMo pointing. Compared to prior VLMs, CoME-VL demonstrates more precise coordinate-level grounding for fine-grained visual queries.}
    \label{fig:qualitative_analysis}
\end{figure*}

\begin{figure*}[t]
    \centering
    \includegraphics[width=\linewidth]{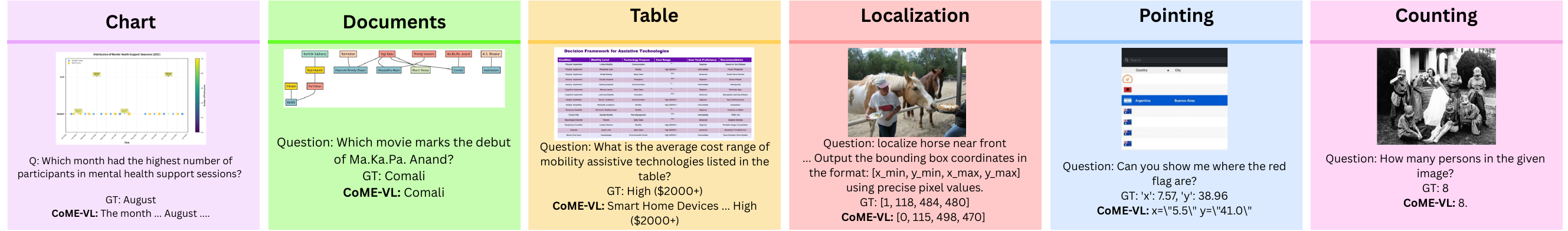}
    \caption{Qualitative examples of CoME-VL on chart understanding, document/table reasoning, localization, pointing, and counting, demonstrating its ability to jointly support visual understanding and grounding.}
    \label{fig:support_tasks}
\end{figure*}

\section{Experiments}\label{sec:experiments}
\subsection{Experimetal Setup} 
We build on Molmo~\cite{deitke2025molmo} with Qwen2-7B (3584-dim, 28 layers) as the language backbone. Vision encoding uses SigLIP-2-SO400M ($384{\times}384$, patch 16, 27 layers) for semantic features and frozen DINOv3-Large ($224{\times}224$, 24 layers) for spatial features. Features are fused via 4D Cross-RoPE attention with $2{\times}2$ pooling and MLP projection. Images are processed at native resolution with overlap-and-resize cropping (max 12 crops). Across all settings, we freeze DINO parameters and update SigLIP parameters during training.

Training uses AdamW ($\beta{=}(0.9,0.95)$, weight decay $0.01$) with a base learning rate of $10^{-4}$ and $10^{-5}$ for the LLM, SigLIP, and connector modules. A cosine learning-rate schedule with 100 warmup steps is applied. Training is conducted with a global batch size of 24, sequence length 2304, BF16 precision, and Fully Sharded Data Parallel (FSDP). The model is trained using \textbf{64$\times$AMD Instinct MI210} GPUs (64 GB each) for 7 days. The training data consists of the Pixmo dataset \cite{deitke2025molmo} and RefCOCO~\cite{mao2016refcocog}. Additional implementation details are provided in Appendix~\ref{appendix:method}, including the Orthogonal Layer design, its optimization parameterization, the redundancy metric, and the bounding-box generation setup used for RefCOCO.

We use the Molmo-defined accuracy metrics, including F1-cap; for Pixmo pointing, we report 3-pixel and 5-pixel counting accuracy, where a prediction is considered correct if its distance to the ground truth is within 3 px or 5 px, respectively, and accuracy is computed as the ratio of correct predictions to the total; RefCOCO evaluation follows the original experimental setup~\cite{kazemzadeh2014refcoco}.

\subsection{Quantitative Analysis}
\paragraph{\textbf{Comparison with State of the Art VLMs.}}
In Table~\ref{tab:single_enc_comp}, we compare our method with state-of-the-art vision--language models (VLMs) on grounding. Since our evaluation is based on the Molmo baseline, which is built upon Qwen2-VL, we restrict comparisons to VLMs released on or before Qwen2-VL to ensure a fair comparison in visual understanding and text generation on the PixMo dataset.

Table~\ref{tab:main_table} reports results on the PixMo benchmarks. Our model achieves 57.24\% on Charts, 66.94\% on Diagrams, 70.75\% on Tables, and 81.84\% on Others, outperforming LLaVA variants. On the Counting task, CoME-VL attains 87.83\%, exceeding InternVL2-8B (74.05\%) and Qwen2-VL-7B (57.42\%). Notably, CoME-VL is the only one reporting Pointing metrics, achieving 58.56\% and 75.94\% accuracy under @3px and @5px thresholds, respectively.

On RefCOCO (Table~\ref{tab:refcoco_singlecol}), our method achieves 92.57\% on the validation set, 95.36\% on testA, and 90.51\% on testB, outperforming both Qwen2-VL and Clip-to-DINO baselines.

\paragraph{\textbf{Comparison with feature-merging methods.}}
Table~\ref{tab:refcoco_singlecol} compares our method against the state-of-the-art feature-merging approach CLIP-to-DINO~\cite{jiang2023clip} and our VLM baselines on RefCOCO~\cite{kazemzadeh2014refcoco} using the original \textit{val}, \textit{testA}, and \textit{testB} splits.
We report the percentage of correct predictions (\%), where a prediction is correct if IoU $\geq$ 0.5.
Our method achieves the best performance across all splits, improving over CLIP-to-DINO by +0.84 (val), +1.30 (testA), and +1.66 (testB).

\begin{table}[h]
\centering
\caption{Performance comparison on the RefCOCO benchmark. Results are reported as localization accuracy (\%) on the val, testA, and testB splits.}
\label{tab:refcoco_singlecol}
\small
\setlength{\tabcolsep}{6pt}
\renewcommand{\arraystretch}{1.2}

\begin{tabular}{lccc}
\toprule
\textbf{RefCOCO} & \textbf{val} & \textbf{testA} & \textbf{testB} \\
\midrule
Molmo~\cite{deitke2025molmo}       & 0.10    & 0.27    & 0.27    \\
Clip-to-DINO~\cite{jiang2023clip} & 91.73 & 94.06 & 88.85 \\
Qwen-VL~\cite{Qwen-VL}      & 89.36 & 92.23 & 85.36 \\
\rowcolor{ModelGreen} \textbf{CoME-VL }        & \textbf{92.57} & \textbf{95.36} & \textbf{90.51} \\
\bottomrule
\end{tabular}
\end{table}

\subsection{Qualitative Analysis}
Figure~\ref{fig:qualitative_analysis} compares model responses on a PixMo pointing example. For the query \emph{“Locate a clear nose trim”}, Qwen3-VL and LLaVA-1.5 generate only descriptive text and fail to produce spatial localization. Molmo outputs a point-based prediction but with noticeable localization error relative to the ground truth (39.88, 56.93). In contrast, our model (CoME-VL) predicts a more accurate point (38.5, 52.8), closer to the ground-truth location, demonstrating superior fine-grained grounding. This example highlights CoME-VL’s ability to jointly perform visual understanding and precise coordinate-level localization, which is essential for PixMo-style pointing tasks.

Figure~\ref{fig:support_tasks} presents representative examples of the tasks supported by CoME-VL, including chart understanding, document understanding, table reasoning, localization, pointing, and counting. These examples highlight the distinct roles of the two visual encoders in our framework: the SigLIP encoder primarily strengthens the understanding component by providing rich semantic representations, while the complementary DINO encoder enhances the grounding component through stronger spatial and localization cues. Together, they allow CoME-VL to unify semantic understanding and precise visual grounding within a single architecture.

\begin{table}[ht]
\centering
\caption{Effect of SigLIP and DINO layer-range selection on PixMo benchmark performance. Combining deeper SigLIP layers with mid-to-high DINO layers yields the strongest results across tasks.}
\small
\setlength{\tabcolsep}{5pt}
\renewcommand{\arraystretch}{1.0}

\resizebox{0.48\textwidth}{!}{%
\begin{tabular}{
>{\raggedright\arraybackslash}p{2.0cm}
@{\hspace{6pt}\vrule\hspace{6pt}}
ccccccc
}
\toprule
\textbf{Model} &
\textbf{Chart} &
\textbf{Diagrams} &
\textbf{Tables} &
\textbf{Others} &
\textbf{Counting} &
\textbf{Pointing (@3/5px} &
\textbf{Avg. Time}\\
\midrule

Molmo\newline Original
& 52.39 & 62.41 & 66.25 & 76.26 & 83.31 & 53.79 / 68.94 & 1.26s\\

Siglip(0--22) +\newline Dino(0--9)
& 56.17 & 66.12 & 69.86 & 81.12 & 86.97 & 56.68 / 74.59 & 1.37s\\

Siglip(22--27) +\newline Dino(0--9)
& 54.96 & 63.72 & 68.28 & 77.16 & 84.23 & 52.41 / 67.65 & 1.33s\\

Siglip(0--22) +\newline Dino(10--23)
& 56.91 & 66.46 & 70.02 & 81.46 & 87.67 & 57.22 / 75.13 & 1.40s\\

Siglip(22--27) +\newline Dino(10--23)
& 56.06 & 65.98 & 69.72 & 80.88 & 87.21 & 56.95 / 74.87 & 1.34s\\

\rowcolor{ModelGreen} Siglip(0--27) +\newline Dino(10--23)
& 57.24 & 66.94 & 70.75 & 81.84 & 87.83 & 58.56 / 75.94 & 1.52s\\

\bottomrule
\end{tabular}%
}
\label{tab:siglip_dino_ablation}
\end{table}

\subsection{Ablation Studies}

\subsubsection{\textbf{Impact of different DINO variants.}}
In Table~\ref{tab:pixmo_singlecol}, we evaluate CoME-VL with different self-supervised backbones by replacing DINOv3 with earlier DINO variants. Results on PixMo show consistent improvements as the self-supervised encoder progresses from DINOv1 to DINOv2 and DINOv3, with CoME-VL achieving the best performance across both understanding and grounding tasks. This highlights the importance of stronger self-supervised spatial representations.

\begin{table}[t]
\centering
\caption{Performance comparison on the Pixmo benchmark~\cite{deitke2025molmo}. Results are reported as accuracy (\%).}
\label{tab:pixmo_singlecol}
\small
\setlength{\tabcolsep}{3pt}
\renewcommand{\arraystretch}{1.15}

\resizebox{\columnwidth}{!}{%
\begin{tabular}{l@{\hspace{4pt}\vrule\hspace{4pt}}ccccc}
\toprule
\textbf{Model} &
\textbf{Chart} &
\textbf{Diagrams} &
\textbf{Tables} &
\textbf{Others} &
\textbf{Counting} \\
\midrule
CoME-DinoV1~\cite{caron2021emerging_dinov1} & 54.18 & 64.20 & 68.41 & 78.00 & 85.91 \\
CoME-DinoV2~\cite{oquab2023dinov2}          & 55.68 & 65.98 & 69.12 & 78.93 & 86.12 \\
\rowcolor{ModelGreen} CoME-Dino3~\cite{simeoni2025dinov3} & \textbf{57.24} & \textbf{66.94} & \textbf{70.75} & \textbf{81.84} & \textbf{87.83} \\
\bottomrule
\end{tabular}%
}
\end{table}

\subsubsection{\textbf{Effect of the modification made on Molmo.}}
Figure~\ref{fig:qualitative_analysis} reports a component-wise ablation on Molmo. Adding RoPE alignment consistently improves localization over the baseline, and introducing OL yields an additional gain by promoting complementary feature fusion. Combining RoPE and OL achieves the best performance, indicating that spatial alignment and redundancy reduction are complementary.
In this analysis, we use outputs from all layers without entropy-based layer selection. After establishing the importance of orthogonality and RoPE-based cross-attention, we further conduct experiments with layer-wise selection.

\begin{figure}[ht]
    \centering
    \includegraphics[width=\linewidth]{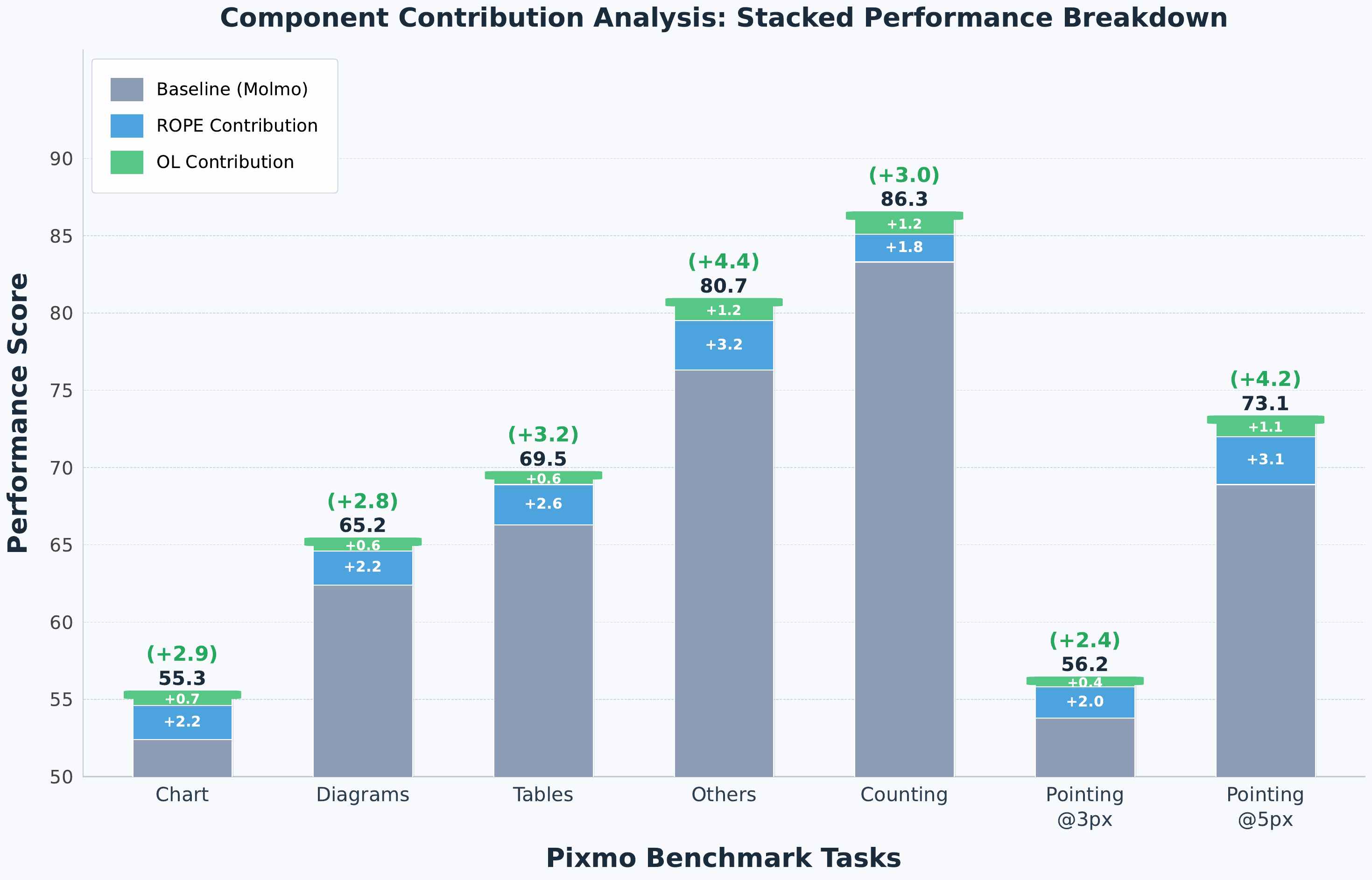}
    \caption{Component-wise contribution analysis on PixMo benchmarks. The stacked bars show the Molmo baseline (gray) and the additive gains from RoPE-based alignment (blue) and orthogonality-regularized fusion (green).}

    \label{fig:qualitative_analysis}
    
\end{figure}

\subsubsection{\textbf{Effect of Multi-scale Features.}}
\label{msf}
Table~\ref{tab:siglip_dino_ablation} analyzes the impact of extracting features from different layer ranges of SigLIP and DINO backbones. Using deeper SigLIP layers (0--27) combined with higher-level DINO layers (10--23) yields the best performance across all PixMo benchmarks, indicating the benefit of integrating both mid- and high-level visual semantics. In contrast, restricting SigLIP to only high layers (22--27) degrades performance, particularly on Diagrams, Counting, and Pointing, suggesting that lower- and mid-level spatial features are critical for precise localization. These results demonstrate that multi-scale feature aggregation from complementary layer ranges is essential for robust visual grounding and counting tasks.








\section{Conclusion}

We presented \textbf{CoME-VL}, a complementary multi-encoder vision-language framework that improves both visual understanding and grounding by jointly leveraging contrastively trained and self-supervised visual representations. Motivated by the distinct semantic and spatial biases of modern vision encoders, CoME-VL combines SigLIP and DINO through entropy-guided layer selection, orthogonality-regularized multi-layer fusion, and RoPE-enhanced cross-attention alignment. Since multi-encoder fusion can introduce redundant features, we impose orthogonality across projected layer representations to encourage complementary feature learning. Furthermore, rather than directly concatenating patches from different encoders, which substantially increases the number of visual tokens processed by the LLM, we adopt RoPE-enhanced cross-attention to align and merge features efficiently.
Our analysis shows that entropy decreases with depth in both encoders, with DINO dropping earlier and more sharply than SigLIP. SigLIP mainly contributes strong semantic features for visual understanding, while DINO provides richer grounding features, and the best results are achieved by combining DINO’s late-layer features with SigLIP features across layers. This highlights their complementary roles and shows that effective multi-encoder fusion can improve performance without adding much burden to the LLM.

\textbf{Limitation:} As shown in Table \ref{tab:siglip_dino_ablation}, CoME-VL incurs a small inference overhead compared with the Molmo baseline, increasing the average time from 1.26s to 1.52s per sample. However, this overhead remains lower than COMM \cite{jiang2023clip}, which requires about 2.2s/sample when CLIP and DINO features are directly concatenated, since the larger fused token set increases the number of visual patches that must be processed by the LLM.

\bibliographystyle{ACM-Reference-Format}
\bibliography{sample-base}

\clearpage
\appendix
\section{Methods Details}
\label{appendix:method}
Given an image $I$ and a referring text query $T$, our goal is to predict the bounding box $b = [x_{\min}, y_{\min}, x_{\max}, y_{\max}]$ that localizes the region in $I$ described by $T$. We formulate this as an autoregressive generation task where the language model produces bounding box coordinates as specialized tokens.


\subsection{Orthogonal Layer (OL)}\label{append:OL}
Multi-layer fusion can be redundant because intermediate representations in deep encoders are often highly correlated across depth. Our goal is to (i) preserve useful information from multiple depths while (ii) reducing overlap among the directions contributed by different layers. To this end, we introduce an \emph{Orthogonal Layer} (OL): a lightweight linear projection applied to each selected layer output before mixing. OL is constrained to be (semi-)orthogonal so that it acts as a near-isometry (no degenerate scaling) and encourages the mixed features to occupy complementary subspaces.

Let encoder $e$ produce a sequence of hidden states at layer $l$:
$\mathbf{Z}_{e}^{(l)} \in \mathbb{R}^{T \times d}$, where $T$ is the number of tokens and $d$ is the hidden dimension.
We select a subset of layers $\mathcal{S}_e$ for mixing (e.g., evenly spaced layers). For each selected layer $l \in \mathcal{S}_e$, we apply LayerNorm and then an orthogonal projection to obtain $\widehat{\mathbf{Z}}_{e}^{(l)} \in \mathbb{R}^{T \times m}$, where $m$ is the mixing dimension (often $m=d$ in our implementation).

\subsection{Layer aggregation (mixing)}
After OL, we mix the selected layer features with learned nonnegative weights:
\begin{equation}
\mathbf{Z}_e = \sum_{l\in\mathcal{S}_e}\alpha_{e,l}\,\widehat{\mathbf{Z}}_{e}^{(l)},
\qquad
\alpha_{e,l}=\frac{\exp(s_{e,l})}{\sum_{k\in\mathcal{S}_e}\exp(s_{e,k})},
\label{eq:layer_mix}
\end{equation}
where $s_{e,l}$ are learned scalars (one per selected layer). The softmax constraint ensures stability and interpretability (weights sum to $1$).
We then use $\mathbf{Z}_e$ as the mixed representation for downstream modules.


\subsection{Bounding-box encoding as tokens}\label{app:box_tokens}
Molmo~\cite{deitke2025molmo} does not natively support detection outputs. To enable autoregressive box prediction, we serialize each bounding box into a short sequence of discrete coordinate tokens.

\paragraph{\textbf{Box parameterization and ordering.}}
Each box is represented in \emph{corner form} as $(x_1,y_1,x_2,y_2)$ with $(x_1,y_1)$ the top-left and $(x_2,y_2)$ the bottom-right corner.
We enforce $x_1\le x_2$ and $y_1\le y_2$ by construction during data preprocessing and decoding (clamping if needed).
The token order is fixed as:
\begin{equation}
\langle \texttt{BOX} \rangle\;\; \langle x_1 \rangle\;\; \langle y_1 \rangle\;\; \langle x_2 \rangle\;\; \langle y_2 \rangle\;\; \langle \texttt{END\_BOX} \rangle.
\label{eq:box_token_order}
\end{equation}

\paragraph{\textbf{Normalization.}}
Coordinates are normalized to $[0,1]$ with respect to the input image size $(W,H)$ used for training/evaluation:
\begin{equation}
\hat{x} = \frac{x}{W},\qquad \hat{y}=\frac{y}{H},
\end{equation}
so each $\hat{x},\hat{y}\in[0,1]$.

\paragraph{\textbf{Discretization (bins) and vocabulary.}}
We quantize each normalized coordinate into $B$ uniform bins (we use $B=1000$ unless otherwise stated):
\begin{equation}
q(\hat{u}) = \left\lfloor \mathrm{clip}(\hat{u},0,1)\cdot (B-1) + 0.5 \right\rfloor \in \{0,\ldots,B-1\}.
\label{eq:quantize}
\end{equation}
Each integer bin index is mapped to a dedicated token from a reserved coordinate vocabulary
$\{\langle \texttt{COORD\_0}\rangle,\ldots,\langle \texttt{COORD\_}(B-1)\rangle\}$.
Thus $\langle x_1\rangle=\langle \texttt{COORD\_}q(\hat{x}_1)\rangle$ and similarly for $y_1,x_2,y_2$.

\paragraph{\textbf{Decoding back to pixels.}}
Given predicted bin indices $k\in\{0,\ldots,B-1\}$, we de-quantize to normalized coordinates
\begin{equation}
\tilde{u}=\frac{k}{B-1},
\end{equation}
and recover pixel coordinates as $\tilde{x}=\tilde{u}W$ and $\tilde{y}=\tilde{u}H$ (followed by clipping to image bounds).
If the model outputs invalid corner ordering (rare), we swap endpoints to ensure $x_1\le x_2$ and $y_1\le y_2$.

\paragraph{\textbf{Autoregressive factorization.}}
With the above serialization, the probability of a single box token sequence factorizes autoregressively as


\begin{multline}
p(\mathbf{y}_{\text{box}} \mid \mathbf{x}) = \\
p(\texttt{BOX}\mid \mathbf{x})
\prod_{t\in\{x_1,y_1,x_2,y_2\}}
p\!\left(\langle t\rangle \mid \mathbf{x}, \texttt{BOX}, 
\langle x_1\rangle,\ldots\right)
\cdot p(\texttt{END\_BOX}\mid \cdot)
\end{multline}

which corresponds to Eq.~(9) in the appendix.

\section{Fusion and Alignment in Multi-Encoder VLMs}\label{app:fusion_alignment}

A common design in recent MLLMs (e.g., LLaVA~\cite{liu2023llava} and Qwen-VL~\cite{Qwen-VL}) is to concatenate visual tokens with text tokens and rely on self-attention in the language backbone for fusion. While effective, this approach increases the context length processed by the LLM, leading to higher compute and memory cost, and can dilute fine-grained visual evidence when the number of visual tokens is large. In contrast, cross-attention-based interfaces (e.g., BLIP-2~\cite{li2023blip}) provide a compact and controllable mechanism by allowing language or visual queries to attend to a smaller set of visual tokens without inflating LLM context length. Below we summarize the trade-offs and provide additional analysis supporting our design choices.

\paragraph{\textbf{Concatenation vs.\ cross-attention fusion.}}
Let $\mathbf{T}\in\mathbb{R}^{N_t\times d}$ denote text tokens and $\mathbf{V}\in\mathbb{R}^{N_v\times d}$ denote visual tokens (after projection to the LLM width).
\textbf{Concatenation fusion} forms $\mathbf{X}=[\mathbf{T};\mathbf{V}]\in\mathbb{R}^{(N_t+N_v)\times d}$ and applies self-attention over all tokens. The per-layer attention cost scales as $\mathcal{O}((N_t+N_v)^2)$ and memory similarly scales with the square of the total sequence length.
\textbf{Cross-attention fusion} keeps the LLM sequence length fixed at $N_t$ and injects visual information via attention from text queries to a (possibly compact) visual memory:
\begin{equation}
\mathrm{CrossAttn}(\mathbf{T},\mathbf{V})=\mathrm{softmax}\!\left(\frac{\mathbf{T}\mathbf{W}_Q(\mathbf{V}\mathbf{W}_K)^\top}{\sqrt{d}}\right)\mathbf{V}\mathbf{W}_V,
\end{equation}
with complexity $\mathcal{O}(N_tN_v)$ per cross-attention layer. When $N_v$ is controlled (e.g., pooled tokens or a learned resampler), cross-attention yields substantially lower cost than concatenation and avoids increasing the LLM context length. This is especially relevant for multi-encoder settings, where concatenation would otherwise add multiple visual token streams.

\paragraph{\textbf{Why alignment matters in multi-encoder fusion.}}
When fusing heterogeneous vision encoders (e.g., contrastive SigLIP-like and self-supervised DINO-like backbones), their token grids may differ in resolution, receptive fields, and preprocessing. Direct concatenation ignores correspondences between tokens and can force the LLM to implicitly learn alignment through self-attention, which is inefficient and brittle under resolution changes. We therefore introduce an explicit alignment stage prior to LLM fusion: a vision-to-vision cross-attention connector that maps one encoder’s tokens into the coordinate frame of the other, producing an aligned and compact visual memory.

\paragraph{\textbf{Geometry-aware alignment and token efficiency.}}
Our RGCA connector aligns SigLIP queries to DINO keys/values and incorporates 2D RoPE so attention logits depend on relative spatial offsets. This design encourages spatially consistent matches across grids while keeping the output token count fixed (equal to the SigLIP token count). Optionally, we pool DINO tokens (e.g., $(2\times2)$ pooling) before cross-attention to further reduce $N_v$ and improve robustness to patch-size mismatch. The resulting fused visual tokens are then projected and passed to the LLM, achieving efficient fusion without increasing the LLM context length.

\section{Semantic Feature Analysis Per Layer}
\label{apd:sementic}
We present a qualitative, layer-wise analysis of semantic and spatial features learned by SigLIP2 and DINOv3 using attention rollout visualizations. Instead of aggregating attention across layers or applying gradient-based weighting, we compute attention rollout independently for each transformer layer, allowing direct inspection of how visual focus evolves with depth.

\subsection{Layer-wise Attention Rollout Visualization}
To visualize how attention evolves across depth, we extract the self-attention matrix from each transformer layer and fuse the attention heads using mean aggregation. We then prune low-importance attention links with a fixed discard ratio while preserving the class-token connections, and recursively multiply the resulting attention matrices to propagate token-to-token relevance within each layer.

\begin{figure*}
    \centering
    \includegraphics[width=0.8\linewidth]{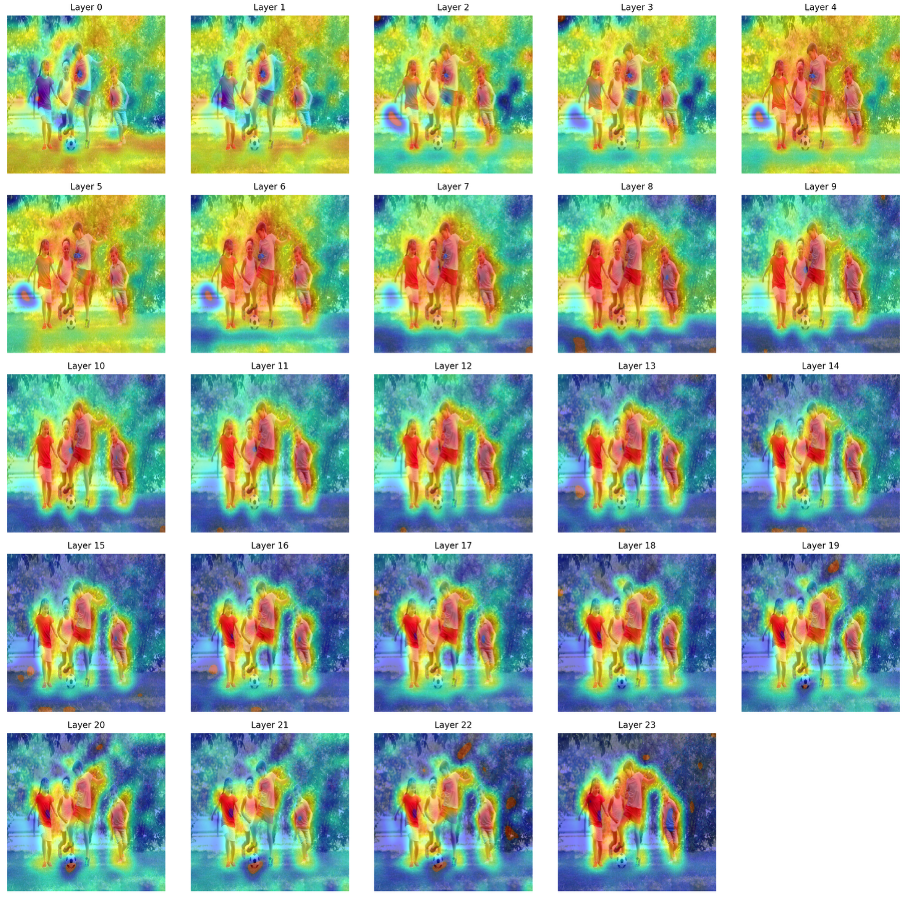}
    \caption{Layer-wise attention rollout for deeper layers: DINOv3 illustrating the transition from spatially coherent object-level attention to semantically discriminative region focus.}
    \label{fig:qualitative_analysis1_a}
\end{figure*}

\begin{figure*}
    \centering
    \includegraphics[width=0.8\linewidth]{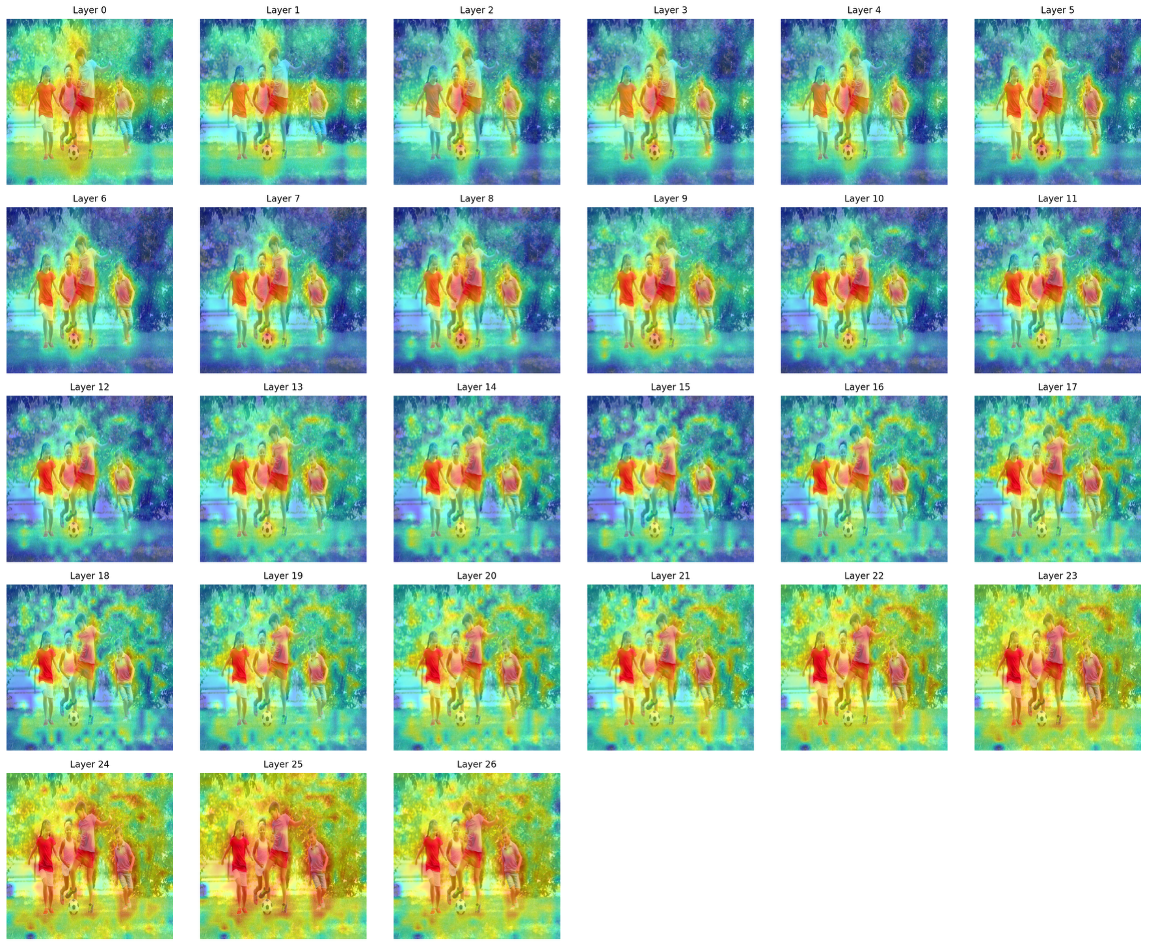}
    \caption{Layer-wise attention rollout for deeper layers: SigLIP2 illustrating the transition from spatially coherent object-level attention to semantically discriminative region focus.}
    \label{fig:qualitative_analysis1_b}
\end{figure*}

\begin{figure*}[ht]
    \centering
    \includegraphics[width=0.8\linewidth]{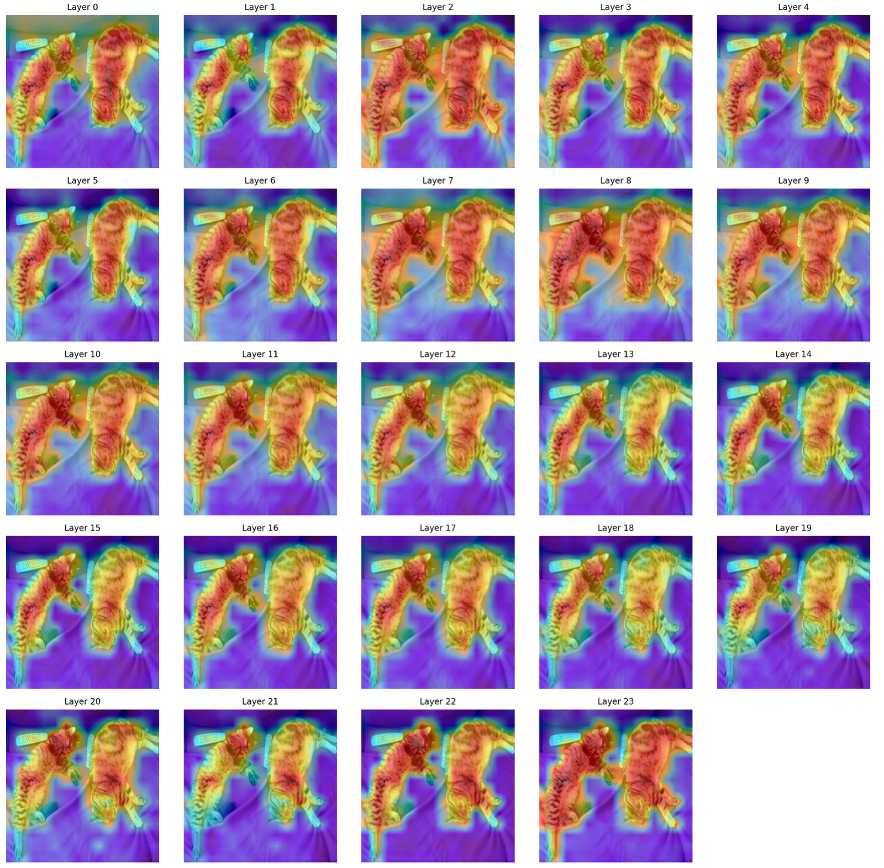}
    \caption{Layer-wise attention rollout for deeper layers: DINOv3 illustrating the transition from spatially coherent object-level attention to semantically discriminative region focus.}
    \label{fig:qualitative_analysis2_a}
\end{figure*}

\begin{figure*}[ht]
    \centering
    \includegraphics[width=0.8\linewidth]{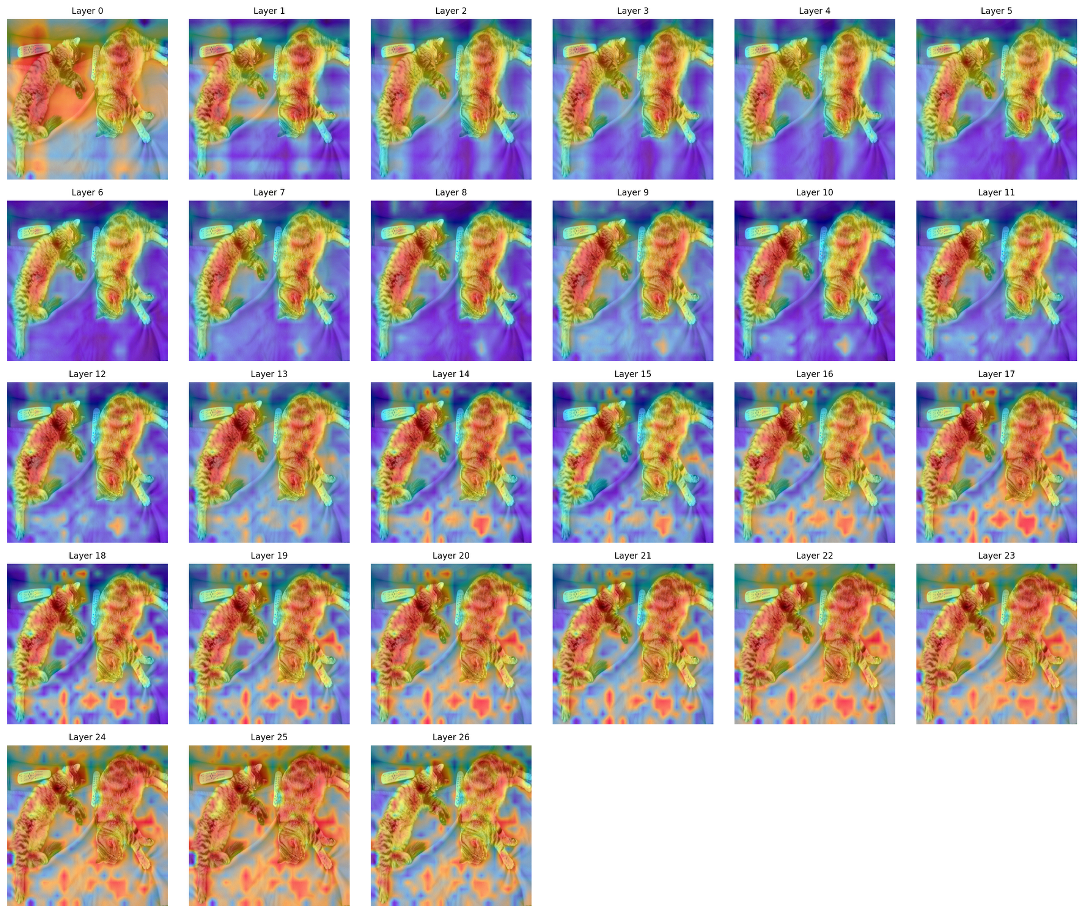}
    \caption{Layer-wise attention rollout for deeper layers: SigLIP2, illustrating the transition from spatially coherent object-level attention to semantically discriminative region focus.}
    \label{fig:qualitative_analysis2_b}
\end{figure*}

The final class-to-patch attention map is reshaped to the spatial grid, normalized, and overlaid on the input image as a heatmap. Repeating this process independently for each layer produces layer-specific visualizations without cross-layer averaging or gradient-based weighting. Figures~\ref{fig:qualitative_analysis1_a}, \ref{fig:qualitative_analysis1_b}, \ref{fig:qualitative_analysis2_a}, and \ref{fig:qualitative_analysis2_b} present the rollout results for deeper DINOv3 and SigLIP2 layers, highlighting their distinct spatial and semantic behaviors across depth.



\paragraph{\textbf{Layer-wise Analysis.}}
The layer-wise attention rollout reveals a clear complementarity between SigLIP2 and DINOv3. SigLIP2 exhibits richer and more informative responses in its early layers, capturing diverse semantic cues with broad spatial coverage, while DINOv3 concentrates discriminative and spatially coherent information in its deeper layers (layers 10–23). This observation directly supports our design choice in the main model: leveraging early SigLIP2 layers for semantic diversity and later DINOv3 layers for precise spatial grounding enables a balanced representation that benefits fine-grained understanding and localization tasks.

\end{document}